# Re-initialization Free Level Set Evolution via Reaction Diffusion

Kaihua Zhang, Lei Zhang, *Member*, *IEEE*, Huihui Song and David Zhang, *Fellow*, *IEEE*

**Abstract** — This paper presents a novel reaction-diffusion (RD) method for implicit active contours, which is completely free of the costly re-initialization procedure in level set evolution (LSE). A diffusion term is introduced into LSE, resulting in a RD-LSE equation, to which a piecewise constant solution can be derived. In order to have a stable numerical solution of the RD based LSE, we propose a two-step splitting method (TSSM) to iteratively solve the RD-LSE equation: first iterating the LSE equation, and then solving the diffusion equation. The second step regularizes the level set function obtained in the first step to ensure stability, and thus the complex and costly re-initialization procedure is completely eliminated from LSE. By successfully applying diffusion to LSE, the RD-LSE model is stable by means of the simple finite difference method, which is very easy to implement. The proposed RD method can be generalized to solve the LSE for both variational level set method and PDE-based level set method. The RD-LSE method shows very good performance on boundary anti-leakage, and it can be readily extended to high dimensional level set method. The extensive and promising experimental results on synthetic and real images validate the effectiveness of the proposed RD-LSE approach.

**Index Terms** — Level set, reaction-diffusion, active contours, image segmentation, PDE, variational method

K. H. Zhang, L. Zhang, and D. Zhang are with Dept. of Computing, The Hong Kong Polytechnic University, Hong Kong, China. Corresponding author: L. Zhang. Email: cslzhang@comp.polyu.edu.hk. This work is supported by the Hong Kong RGC General Research Fund (PolyU 5375/09E).
H. H. Song is with Dept. of Geography and Resource Management, The Chinese University of Hong Kong, Hong Kong, China.

1. **INTRODUCTION**

In the past two decades, active contour models (ACMs, also called snakes or deformable models) [1] have been widely used in image processing and computer vision applications, especially for image segmentation [5][9][12-13][17-18][30][34][41-42][52-53][59]. The original ACM proposed by Kass *et al.* [1] moves the explicit parametric curves to extract objects in images. However, the parametric ACM has some intrinsic drawbacks, such as its difficulty in handling topological changes and its dependency of parameterization [2]. The level set method later proposed by Osher and Sethian [2] implicitly represents the curve by the zero level of a high dimensional function, and it significantly improves ACM by being free of these drawbacks [2-5].

The level set methods (LSM) can be categorized into partial differential equation (PDE) based ones [8] and variational ones [9]. The level set evolution (LSE) of PDE-based LSM is directly derived from the geometric consideration of the motion equations [6], which can be used to implement most of the parametric ACMs, such as Kass *et al.*'s snakes [1], region competition snakes [12], and geodesic active contours [5], etc. The LSE of variational LSM is derived via minimizing a certain energy functional defined on the level set [9], such as Chan-Vese ACM [18], Vese and Chan's piecewise smoothing ACM [42], local binary fitting ACM [30][41], etc. Moreover, the variational LSM can be easily converted into PDE-based LSM by changing slightly the LSE equation while keeping the final steady state solution unchanged [7].

In implementing the traditional LSMs [4-5][8][10][18], the upwind schemes are often used to keep numerical stability, and the level set function (LSF) is initialized to be a signed distance function (SDF). Since the LSF often becomes very flat or steep near the zero level set in the LSE process and this will affect much the numerical stability [8][14], a remedy procedure called re-initialization is applied periodically to enforce the degraded LSF being an SDF [14]. The first re-initialization method was proposed by Chopp [35] and it directly computes the SDF. However, this method is very time-consuming. In [35], Chopp also proposed a more efficient method by restricting the front movement and the re-initialization within a band of points near the zero level set. However, it is difficult to locate and discretize the interface by Chopp's methods [10]. The method proposed by Sussman *et al.* [14] iteratively solves a re-initialization equation. Nonetheless, when the LSF is far away from an SDF, this method fails to yield a desirable SDF. The re-initialization method in [8] addresses this problem by using a new signed function, but it will shift the interface to some degree [10]. In order to make the interface stationary during re-initialization, a specific method for the two-phase incompressible flow was proposed in [39], which focuses on preserving the amount of material in each cell. The method in [38] uses a true upwind discretization near the interface to make the interface localization accurate, and it can keep the interface stationary. All the above mentioned re-initialization methods, however, have the risk of preventing new zero contours from emerging [16], which may cause undesirable results for image segmentation, such as failures to detecting the interior boundary.

In recent years, some variational level set formulations [9][34][59] have been proposed to regularize the LSF during evolution, and hence the re-initialization procedure can be eliminated. These variational LSMs without re-initialization have many advantages over the traditional methods [4-5][8][10][18], including higher efficiency and easier implementation, etc [9]. Some global minimization methods [61][62] eliminate the re-initialization procedure by combining the total variational model with the Chan-Vese model [18] or the Vese-Chan's piecewise smoothing model [42]. However, these global minimization methods [61][62] can only be applied to some variational LSF with specific forms.

In this paper we propose a new LSM, namely the reaction-diffusion (RD) method, which is completely free of the costly re-initialization procedure. The RD equation was originally used to model the chemical mechanism of animal coats [58]. It includes two processes: *reaction*, in which the substances are transformed into each other, and *diffusion*, which causes the substances to spread out over a surface in space. The RD equation was also used to describe the dynamic process in fields such as texture analysis [55-57], natural image modeling [54] and phase transition modeling [21][25-28][31]. In particular, the RD equation in phase transition modeling is based on the Van der Waals-Cahn-Hilliard theory [26], which is widely used in mechanics for stability analysis of systems with unstable components (e.g., density distributions of a fluid confined to a container [31]). It has been proved that the stable configurations of the components are piecewise constant in the whole domain, and the interfaces between the segmented areas have minimal length [21][26]. These conclusions have been used in image classification with promising results [33]. However, the phase transition method cannot be directly applied to image segmentation because of the inaccurate representation of interface and the stiff parameter $\varepsilon^{-1}$ in its RD equation [16].

The joint use of phase transition and LSM has been briefly discussed in [60]. However, [60] aims to apply the curvature-related flow in phase transition to analyze the evolution driven by the curvature based force. In fact, the curvature motion based on the phase transition theory has been widely studied [11][13][17][48-51]. For example, the classical Merriman-Bence-Osher (MBO) algorithm [11] applies a linear diffusion process to a binary function to generate the mean curvature motion with a small time step; the methods in [13][48-51] convolve a compactly supported Gaussian kernel (or an arbitrary positive radically symmetric kernel) with a binary function to generate similar motion (often called the convolution-generated curvature motion).

Motivated by the RD based phase transition theory [27], we propose to introduce a diffusion term into the conventional LSE equation, constructing a RD-LSE equation to combine the merits of phase transition and LSM. We present the unique and stable equilibrium solution of RD-LSE based on the Van der Waals-Cahn-Hilliard theory, and give the accurate representation of interface. The re-initialization procedure is completely eliminated from the proposed RD-LSM owe to the regularization of the diffusion term. A two-step splitting method (TSSM) is proposed to iteratively solve the RD equation in order to eliminate the side effect of the stiff parameter $\varepsilon^{-1}$. In the first step of TSSM, the LSE equation is iterated, while in the second step the diffusion equation is solved, ensuring the smoothness of the LSF so that the costly re-initialization procedure is not necessary at all. Though the diffusion method has been widely used in image processing, to the best of our knowledge, our work is the first one to apply diffusion to LSE, making it re-initialization free with a solid theoretical analysis under the RD framework. One salient advantage of the proposed method is that it can be generalized to a unified framework whose LSE equation can be either PDE-based ones or variational ones. Another advantage of RD-LSE is its higher boundary anti-leakage and anti-noise capability compared with state-of-the-art methods [9][34][59]. In addition, due to the diffusion term, the LSE formulation in our method can be simply implemented by finite difference scheme instead of the upwind scheme used in traditional LSMs [5][8][10][18]. The proposed RD-LSE method is applied to representative ACMs such as geodesic active contours (GAC) [5] and Chan-Vese (CV) active contours [18]. The results are very promising, validating the effectiveness of RD-LSE.

The rest of the paper is organized as follows. Section 2 introduces the background and related works. Section 3 presents the RD-LSM. Section 4 implements RD-LSM, and analyzes the consistency between theory and implementation. Section 5 presents experimental results and Section 6 concludes the paper.

## 2. BACKGROUND AND RELATED WORKS

### 2.1 Level Set Method

Consider a closed parameterized planar curve or surface, denoted by $C(p,t):[0,1]\times R^+ \to R^n$, where $n=2$ is for planar curve and $n=3$ is for surface, and $t$ is the artificial time generated by the movement of the initial curve or surface $C_0(p)$ in its inward normal direction $\vec{N}$. The curve or surface evolution equation is as follows

$$\begin{cases} C_t = F\vec{N} \\ C(p,t=0) = C_0(p) \end{cases} \quad (1)$$

where $F$ is the force function [10]. For parametric ACMs [1][5][12], we can use the Lagrangian approach to getting the above evolution equation and solve it iteratively. However, the intrinsic drawback of iteratively solving Eq. (1) lies in its difficulty to handle topological changes of the moving front, such as splitting and merging [2]. This problem can be avoided by using the LSM [2]. Consider a closed moving front $C(t)= x \in R^n$, which is represented by the zero level set of an LSF $\phi(x,t)$, i.e., $C(t)=\{x|\phi(C(t),t) = 0\}$. Since $\phi(C(t),t)=0$, we can take the derivative w.r.t time $t$ on both sides, yielding the following equation

$$\begin{cases} \phi_t + \nabla\phi \cdot C_t = \phi_t + \nabla\phi \cdot F\vec{N} = 0 \\ \phi(x,t=0) = \phi_0(x) \end{cases} \quad (2)$$

where gradient operator $\nabla(\cdot) \triangleq (\partial(\cdot)/\partial x_1, \partial(\cdot)/\partial x_2,\ldots,\partial(\cdot)/\partial x_n)$, and $\phi_0(x)$ is the initial LSF $C_0(p)=\{x|\phi_0(x)=0\}$. Since the inward normal can be represented as $\vec{N} = -\nabla\phi/|\nabla\phi|$ [2][4][6], Eq. (2) can be re-written as

$$\begin{cases} \phi_t = F|\nabla\phi| \\ \phi(x,t=0) = \phi_0(x) \end{cases} \quad (3)$$

In traditional LSMs [4][5][10], $F$ is often written as $F=\alpha\kappa+F_1$, where $\kappa=\text{div}(\nabla\phi/|\nabla\phi|)$ is the curvature (the divergence operator is defined as $\text{div}(\vec{V}) \triangleq \sum_{i=1}^{n} \partial v_i/\partial x_i$ with $\vec{V} = (v_1,v_2,\ldots,v_n)$), $\alpha$ is a fixed parameter, and the remaining term $F_1$ can be a constant [4][10].

Eq. (3) is the LSE equation of PDE-based LSMs [2][5][10][13], and it is derived from the geometric consideration of the motion equation [6]. In variational LSMs [7][9][18][30][41-42], the LSE equation is

$$\begin{cases} \phi_t = -E_\phi(\phi) = F\delta(\phi) \\ \phi(x,t=0) = \phi_0(x) \end{cases} \quad (4)$$

where $E_\phi(\phi)$ denotes the Gateaux derivative (or first variation) [22] of an energy functional $E(\phi)$, $\delta(\phi)$ is the Dirac functional, and $F$ has the form as defined in Eq. (3).

In implementing traditional LSMs [4-5][8][10][18], the term $F_1$ in the force function $F$ of Eq. (3) or Eq. (4) is usually approximated by using the upwind scheme, while the remaining can be approximated by the simple central difference scheme [4][19]. During evolution, the LSF may become too flat or too steep near the zero level set, causing serious numerical errors. Therefore, a procedure called re-initialization [7-8][10][14] is periodically employed to reshape it to be an SDF.

**2.2 Re-initialization [7-8][10][14] vs. Without Re-initialization [9][34][59]**

*A. Re-initialization:* In [2], Osher and Sethian proposed to initialize the LSF as $\phi(x) =1\pm dist^2(x)$, where $dist(\cdot)$ is a distance function and "±" denotes the signs inside and outside the contour. Later, Mulder *et al.* [36] initialized the LSF as $\phi(x)=\pm dist(x)$, which is an SDF that can result in accurate numerical solutions. However, in evolution the LSF can become too steep or flat near the contour, leading to serious numerical errors. In order to reduce numerical errors, Chopp [35] periodically re-initialized the LSF to be an SDF. Unfortunately, this re-initialization method straightforwardly computes the SDF in the whole domain and it is very time-consuming. Chopp also proposed [35] to restrict the re-initialization to a band of points close to the zero level set. Such a narrow band method [8][15] can reduce the computational complexity to some extent.

Many lately developed re-initialization methods do not directly compute the SDF [7-8][10][14] since the solution of $|\nabla\phi|=1$ is itself an SDF [37]. In [14], the following re-initialization equation was proposed

$$\phi_t + S(\phi_0)(|\nabla\phi|-1) = 0 \tag{5}$$

where $S(\phi_0) \triangleq \phi_0/\sqrt{\phi_0^2+(\Delta x)^2}$, $\phi_0$ is the initial LSF and $\Delta x$ is the spatial step. Unfortunately, if the initial LSF $\phi_0$ deviates much from an SDF, Eq. (5) will fail to yield a desirable final SDF [8][38]. This problem can be alleviated by modifying $S(\cdot)$ as $S(\phi) \triangleq \phi/\sqrt{\phi^2+|\nabla\phi|^2(\Delta x)^2}$ [8]. However, this method will shift the interface from its original position [10]. Some methods were then proposed to make the interface remain stationary during re-initialization. For example, [39] is specific to the two-phase incompressible flow by preserving the amount of material in each cell, while [38] uses a true upwind discretization in the neighborhood of the interface to achieve the accurate interface localization and keep the interface stationary.

In summary, re-initialization has many problems, such as the expensive computational cost, blocking the emerging of new contours [17], failures when the LSF deviates much from an SDF, and inconsistency between theory and implementation [40]. Therefore, some formulations have been proposed to regularize the variational LSF to eliminate the re-initialization procedure [9][59][34].

***B. Distance regularized level set evolution (DRLSE) [9][59][34]:*** In [9], Li *et al.* proposed a signed distance penalizing energy functional:

$$P(\phi) = \frac{1}{2}\int_\Omega (|\nabla\phi|-1)^2 d\mathrm{x} \tag{6}$$

Eq. (6) measures the closeness between an LSF $\phi$ and an SDF in the domain $\Omega \subset R^n$, $n$=2 or 3. By calculus of variation [22], the gradient flow of $P(\phi)$ is obtained as

$$\phi_t = -P_\phi(\phi) = \mathrm{div}[r_1(\phi)\nabla\phi] \tag{7}$$

Eq. (7) is a diffusion equation with rate $r_1(\phi)=1-1/|\nabla\phi|$. However, $r_1(\phi)\to-\infty$ when $|\nabla\phi|\to 0$, which may cause oscillation in the final LSF $\phi$ [59]. In [59], this problem is solved by applying a new diffusion rate

$$r_2(\phi) = \begin{cases} \dfrac{\sin(2\pi|\nabla\phi|)}{2\pi|\nabla\phi|}, & \text{if } |\nabla\phi|\leq 1 \\ 1-\dfrac{1}{|\nabla\phi|}, & \text{if } |\nabla\phi|\geq 1 \end{cases} \tag{8}$$

Xie [34] also proposed a constrained level set diffusion rate as

$$r_3(\phi) = \mathcal{H}_\rho(|\nabla\phi| - 1) \tag{9}$$

where $\mathcal{H}_\rho(z) = (1/2)[1+(2/\pi)\arctan(z/\rho)]$ and $\rho$ is a fixed parameter.

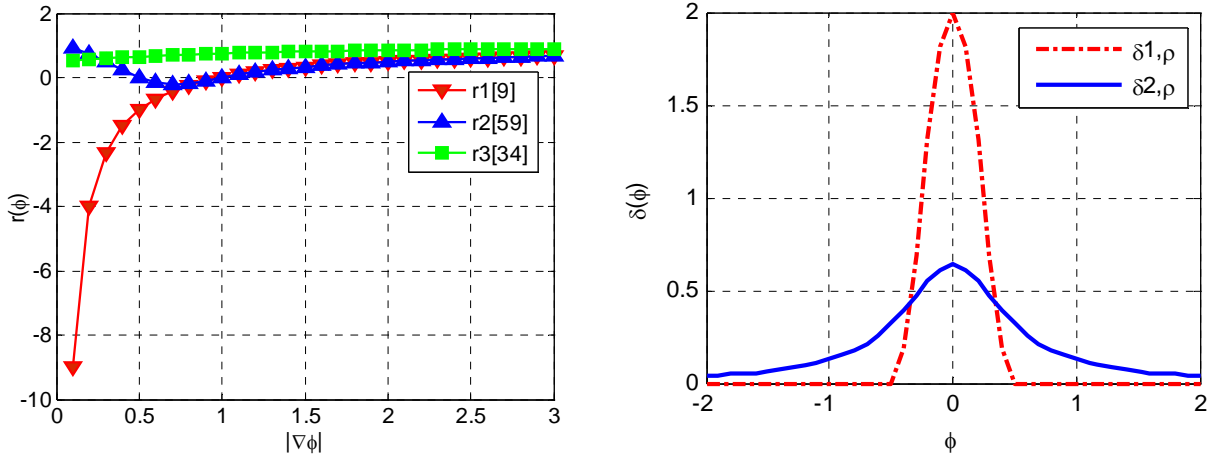

Fig. 1: Left: different diffusion rates; Right: profiles of the two commonly used Dirac functional $\delta_{1,\rho}$ and $\delta_{2,\rho}$.

The three diffusion rates are illustrated in the left figure of Fig. 1. We see that the diffusion rate in [9] constrains the LSF to be an SDF, so does the one in [59] when $|\nabla\phi|\geq 0.5$. When $|\nabla\phi|\leq 0.5$, the diffusion rate in [59] makes the LSF flat, preventing the emerging of unnecessary peaks and valleys. The diffusion rate in [34] changes smoothly from 0 to 1 and makes the LSF tend to flat. The diffusion equation in Eq. (7) with different diffusion rates can be combined into Eq. (4), resulting in the following LSE equation:

$$\begin{cases} \phi_t = \text{Reg}(\phi) + F\delta(\phi) \\ \phi(\mathrm{x}, t=0) = \phi_0(\mathrm{x}) \end{cases} \tag{10}$$

where $\text{Reg}(\phi) = \alpha \text{div}(r(\phi)\nabla\phi)$, $r(\phi) = r_1(\phi)$, $r_2(\phi)$ or $r_3(\phi)$, and $\alpha$ is a constant. The Dirac functional $\delta(\phi)$ can be approximated by the following two forms

$$\delta_{1,\rho}(z) = \begin{cases} 0, & z \in R, |z| > \rho \\ \dfrac{1}{2\rho}\left[1 + \cos\left(\dfrac{\pi z}{\rho}\right)\right], & |z| \leq \rho \end{cases} \tag{11}$$

$$\delta_{2,\rho}(z) = \frac{1}{\pi} \cdot \frac{\rho}{\rho^2 + z^2}, z \in R \tag{12}$$

As shown in the right figure of Fig. 1, the support of $\delta_{1,\rho}(z)$ is restricted into a neighborhood of zero level set so that the LSE can only act locally. The evolution is easy to be trapped into local minima. In contrast, $\delta_{2,\rho}(z)$ acts on all level curves, and hence new contours can appear spontaneously, which makes it tend to yield a global minimum [18]. Thus, $\delta_{2,\rho}(z)$ is widely used in many LSMs [17-18][30][34][41-42][52-53].

**Remark 1:** Since the three DRLSE methods [9][59][34] can be generalized into the same formulation defined in Eq. (10) but with different force term *F*, in the following of this paper we call them generalized DRLSE (GDRLSE). The GDRLSE methods using $r_1(\phi)$ [9], $r_2(\phi)$ [59], and $r_3(\phi)$ [34] are called as GDRLSE1, GDRLSE2, and GDRLSE3, respectively. In our experiments in Section 5, *F* will be constructed by constant term and curvature term (Section 5.2), edge-based force term (Section 5.3), GAC force term (Section 5.4), and CV force term (Section 5.5), respectively.

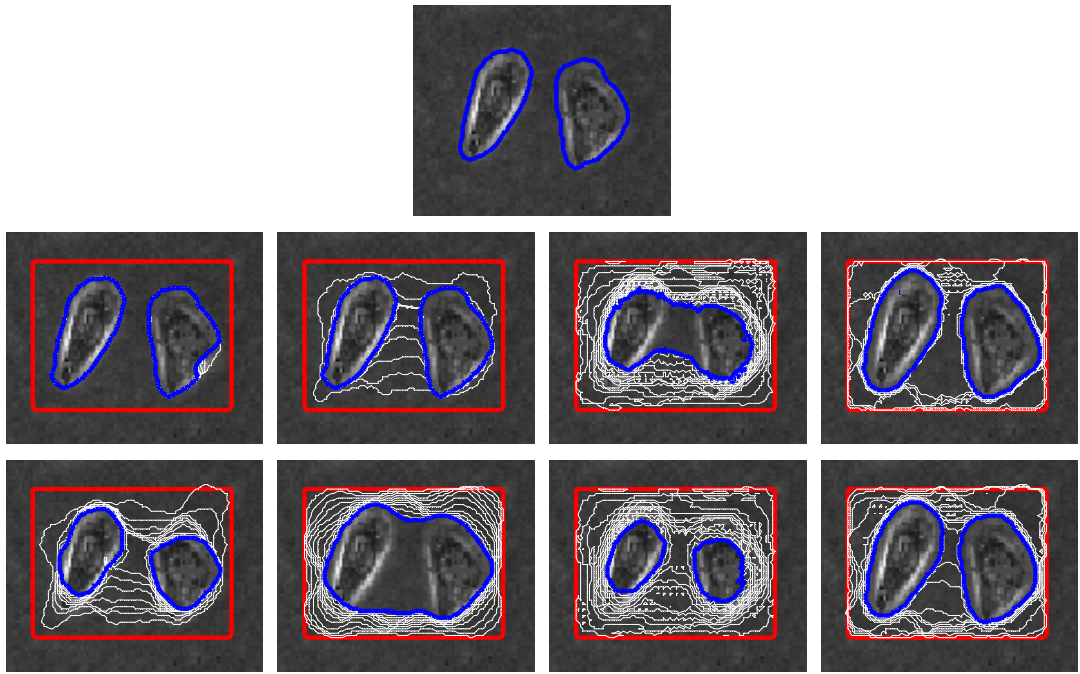

**Fig. 2: An example segmentation of image with weak edges. Top row: ground truth. Middle and bottom rows: from left to right we show the results by GDRLSE1 (the code and test image are downloaded from [29]), GDRLSE2, GDRLSE3 and the proposed RD method with edge-based force term. Middle row: results by using Dirac functional $\delta_{1,\rho}$. Bottom row: results by using Dirac functional $\delta_{2,\rho}$. The red curve represents the initial contour; the blue curve represents the final contour; and the contours during LSE are represented by white solid lines.**

*C. Problems of GDRLSE:* Although GDRLSE methods have many advantages over re-initialization methods, such as higher efficiency and easier implementation, they still have the following drawbacks.

- *Limited application to PDE-based LSMs:* In [59], Li *et al*. claimed that GDRLSE2 can be readily extended to PDE-based LSM. Similarly, GDRLSE1 and GDRLSE3 can also be extended to PDE-based LSM. However, no experiment or theoretical analysis was presented in [59]. We found that these three methods cannot work well for PDE-based LSE. The reason is as follows. When applying GDRLSE methods to PDE-based LSE, we can rewrite Eq. (10) as $\phi_t = F_R|\nabla(\phi)| + F|\nabla(\phi)|$, where $F_R = \alpha\nabla(r(\phi)\nabla\phi)/|\nabla(\phi)|$ is the regularization force that drives zero level set to evolve. When the zero level set reaches the object boundary, the LSE force *F* will be close to zero [5][18], and this can make the zero level set finally stop at the object boundary if the regularization force $F_R=0$ with $r(\phi)=0$ [22], i.e., $|\nabla\phi|=1$ when $r(\phi)=r_1(\phi)$ or $r_2(\phi)$. However, when $r(\phi)=r_3(\phi)$, no solution satisfies $r(\phi)=0$ (please refer to the left image in Fig. 1). Thus, when the zero level set reaches the object boundary, the LSF $\phi$ may not be an SDF (i.e., $|\nabla\phi|\neq 1$), making the regularization force $F_R$ be nonzero, driving the zero level set continue to evolve, and finally causing the boundary leakage problem. Our experimental results in Section 5.4 also validate our above analysis (please refer to Fig. 9 in Section 5.4).

- *Limited anti-leakage capability for weak boundaries:* As can be seen in the middle row of Fig. 2, when using $\delta_{1,\rho}$ and edge-based force term, GDRLSE1 [9], GDRLSE2 [59] and our RD methods all can yield a satisfactory segmentation without boundary leakage. However, the good anti-leakage capability of GDRLSE1 and GDRLSE2 comes not only from their regularization term, but also from the use of Dirac functional $\delta_{1,\rho}$ that restricts the force function to act only in a small neighborhood of zero level set (see the red dotted line in the right figure of Fig. 1). Nonetheless, the use of $\delta_{1,\rho}$ prevents the emerging of new contours, which may make LSE fall into local minima. This is why $\delta_{1,\rho}$ is not used in the CV model [18]. If we use the Dirac functional $\delta_{2,\rho}$, the force function will act on curves at all levels, but this will easily lead to boundary leakage (see the bottom left figure in Fig. 2). The PDE-based LSE can also make the force function act on curves at all levels by replacing $\delta_{1,\rho}$ with $|\nabla\phi|$ [18]. By applying GDRLSE methods to PDE-based LSE, it can also be found that the boundary leakage will occur for objects with weak boundary (refer to Fig. 9 in Section 5.4). The method in [34] uses $\delta_{2,\rho}$ but employs a specific force term so that it can produce good results for images with weak boundaries. If other commonly used force terms are used, however, it can be shown that the boundary leakage will easily occur (see the third figure, bottom row, Fig. 2). The reason of bad performance by using $\delta_{2,\rho}$ in GDRLSE methods is the same as that by using $|\nabla\phi|$ in PDE-based LSE.

- *Sensitivity to noise:* When the image is contaminated by strong noise, the force term $F$ in Eq. (10) will be much distorted, and then the regularization term will fail to keep the LSF smooth. Please refer to Fig. 10 in Section 5.5 for examples.

## 3. REACTION-DIFFUSION (RD) BASED LEVEL SET EVOLUTION

Since the zero level is used to represent the object contour, we only need to consider the zero level set of the LSF. As pointed out in [8], with the same initial zero level set, different embedded LSFs will give the same final stable interface. Therefore, we can use a function with different phase fields as the LSF. Motivated by the phase transition theory [20][27], we propose to construct a RD equation by adding a diffusion term into the conventional LSE equation. Such an introduction of diffusion to LSE will make LSE stable without re-initialization. We will show in **Theorem 2** that the stable solution of the RD equation is piecewise constant with different phase fields in the domain $\Omega$, and it is also the solution of the LSE equation.

By adding a diffusion term "$\varepsilon\Delta\phi$" into the LSE equation in Eq. (3) or Eq. (4), we have the following RD equation for LSM:

$$\begin{cases} \phi_t = \varepsilon\Delta\phi - \frac{1}{\varepsilon}L(\phi), x \in \Omega \subset R^n \\ \text{subject to } \phi(x, t=0, \varepsilon) = \phi_0(x) \end{cases} \quad (13)$$

where $\varepsilon$ is a small positive constant, $L(\phi) = -F|\nabla\phi|$ for PDE-based LSM or $L(\phi) = -F\delta(\phi)$ for variational LSM, $\Delta$ is the Laplacian operator defined by $\Delta(\cdot) \triangleq \sum_{i=1}^{n} \partial^2(\cdot)/\partial x_i^2$, and $\phi_0(x)$ is the initial LSF. Eq. (13) has two dynamic processes: the *diffusion* term "$\varepsilon\Delta\phi$" gradually regularizes the LSF to be piecewise constant in each segment domain $\Omega_i$, and the *reaction* term "$-\varepsilon^{-1}L(\phi)$" forces the final stable solution of Eq. (13) to $L(\phi)=0$, which determines $\Omega_i$. In the traditional LSMs [4-5][8][10][18], due to the absence of the diffusion term we have to regularize the LSF by an extra procedure, i.e., re-initialization.

In the following, based on the Van der Waals-Cahn-Hilliard theory of phase transitions [26], we will first analyze the equilibrium solution of Eq. (13) when $\varepsilon \to 0^+$ for variational LSM, and then generalize the analysis into a unified framework for both PDE-based LSM and variational LSM.

**Theorem 1:** *Let $\Omega \subset R^n$, n=2 or 3, be the domain of the level set function $\phi$ and assume that $E(\phi)$ is an energy functional w.r.t. $\phi$, the Euler equations of $E(\phi)$ and $F(\phi)$ are the same, i.e., $E_\phi(\phi)=F_\phi(\phi)$, where $F(\phi) \triangleq \int_\Omega E(\phi)dx$.* ∎

**Proof**: see **Appendix A** please. ∎

For variational LSM, assuming that the $L(\phi)$ in Eq. (13) is obtained by minimizing an energy functional $E(\phi)$, i.e., $L(\phi)=E_\phi(\phi)$, then according to **Theorem 1** we can obtain the following energy functional $F_\varepsilon(\phi)$ whose gradient flow is Eq. (13):

$$F_\varepsilon(\phi) = \frac{1}{2}\varepsilon\int_\Omega |\nabla\phi|^2 \, dx + \frac{1}{\varepsilon}\int_\Omega E(\phi)dx \quad (14)$$

We can use the Van der Waals-Cahn-Hilliard theory of phase transitions [31] to analyze Eq. (14). Consider a dynamical system composed of a fluid whose Gibbs free energy per volume is prescribed by an energy functional $E(\phi)$ w.r.t. the density distribution $\phi(x): \Omega \to R$, subject to isothermal conditions and confined to a bounded container $\Omega \subset R^n$, which is an open bounded subset of $R^n$ with Lipschitz continuous boundaries [31]. The stable configuration of the fluid is obtained by solving the following variational problem $P_0$ [21][31]:

$$P_0 : \begin{cases} \inf_{\phi} F_0(\phi) = \int_\Omega E(\phi)dx \\ \text{subject to} \quad \int_\Omega \phi(\mathrm{x})dx = m \end{cases} \quad (15)$$

where $m$ is the total mass. It is obvious that $P_0$ is a non-convex problem. The Van der Waals-Cahn-Hilliard theory introduces a simple singular perturbation $\varepsilon|\nabla\phi|^2/2$ with a very small constant $\varepsilon$ to ensure the uniqueness of the solution to $P_0$. The problem of Eq. (15) is then changed to

$$P_\varepsilon : \begin{cases} \inf_{\phi} F_\varepsilon(\phi) \\ \text{subject to} \quad \int_\Omega \phi(\mathrm{x})dx = m \end{cases} \quad (16)$$

where $F_\varepsilon(\phi)$ is defined in Eq. (14). The equilibrium solution of Eq. (13) is then the solution of $P_\varepsilon$ in Eq. (16). The $\Gamma$-convergence theory has been used to study the problem $P_\varepsilon$ as $\varepsilon\to 0^+$ [21][28][31-32].

Consider the LSE equation (13) with $L(\phi)=E_\phi(\phi)$, which is the gradient flow of $F_\varepsilon(\phi)$ in Eq. (14), there exists an important theorem as follows.

**Theorem 2:** *If there are $k\geq 2$ local minima $c_1,\ldots,c_k$ for the energy functional $E(\phi)\geq 0$ in the domain $\Omega$ such that $\{E(c_i)=0, i=1,\ldots,k\}$, then for the point $\mathrm{x}$ where the initial function $\phi_0(\mathrm{x})$ is in the basin of attraction of $c_i$, the solution $\phi(\mathrm{x},t,\varepsilon)$ of $P_\varepsilon$ will approach to $c_i$ as $\varepsilon\to 0^+$, which is also the equilibrium solution of the LSE equation $\phi_t = -E_\phi(\phi)$ with the same initialization $\phi(\mathrm{x},t=0)=\phi_0(\mathrm{x})$, i.e.,*

$$\lim_{t\to+\infty, \varepsilon\to 0^+} \phi(\mathrm{x},t,\varepsilon) = \sum_{i=1}^k c_i \chi_i(\mathrm{x}) \quad (17)$$

*where $\chi_i(\mathrm{x})\in\{0,1\}$ is the characteristic function of set $S_i=\{\mathrm{x}|\phi_0(\mathrm{x})\in B_i, i=1,\ldots,k\}$, and $B_i$ is a basin to attract $\phi(\mathrm{x},t,\varepsilon)$ to $c_i$.* ∎

**Proof**: see **Appendix B** please. ∎

**Theorem 2** can be readily extended to the case when $E_\phi(\phi)$ is replaced by a function $L(\phi)$ that is not the gradient of a potential, where the $k$ local minima $c_1,\ldots,c_k$ of energy functional $E(\phi)$ are replaced by the stable zeros of $L(\phi)$. The proof for **Theorem 2** can be found in **Appendix B** and we have the following remark.

**Remark 2**: For PDE-based LSM, we have $L(\phi)=-F|\nabla\phi|$ in Eq. (13). It is obvious that any constant $c\neq 0$ can make $L(\phi)=0$. As claimed in **Remark A-2** in **Appendix B**, if there exist $k$ zeros for $L(\phi)$ in domain $\Omega$, the final equilibrium solution of Eq. (13) can be represented by $\phi(\mathrm{x},t,\varepsilon)=\sum_{i=1}^k c_i\chi_i(\mathrm{x})$, $\varepsilon\to 0^+$, where $c_i$ is determined by the initial LSF $\phi_0(\mathrm{x})$. Therefore, the above theoretical analysis is applicable to both PDE-based LSM and variational LSM, whose LSE equations can be unified into the RD framework in Eq. (13).

---

**Algorithm 1: RD based level set evolution (RD-LSE)**

---

1. Initialization: $\phi^n = \phi_0$, $n = 0$
2. Compute $\phi^{n+1/2}$ as
$$\phi^{n+1/2} = \phi^n - \Delta t_1 \cdot L(\phi^n) \tag{18}$$
3. Compute $\phi^{n+1}$ as
$$\phi^{n+1} = \phi^n + \Delta t_2 \cdot \Delta \phi^n \tag{19}$$
   where $\phi^n = \phi^{n+1/2}$.
4. If $\phi^{n+1}$ satisfies stationary condition, stop; otherwise, $n = n + 1$ and return to Step 2.

---

## 4. IMPLEMENTATION

From the analysis in Section 3, we see that the equilibrium solution of Eq. (13) is piecewise constant as $\varepsilon \to 0^+$, which is the characteristic of phase transition [20][28]. On the other hand, Eq. (13) has the intrinsic problem of phase transition, i.e., the stiff parameter $\varepsilon^{-1}$ makes Eq. (13) difficult to implement [11][16][44]. In this section, we propose a splitting method to implement Eq. (13) to reduce the side effect of stiff parameter $\varepsilon^{-1}$.

### 4.1 Two-Step Splitting Method (TSSM) for RD

A TSSM algorithm to implement RD has been proposed in [11] to generate the curvature-dependent motion. In [11] the reaction function is first forced to generate a binary function with values 0 and 1, and then the diffusion function is applied to the binary function to generate curvature-dependent motion. Different from [11], where the diffusion function is used to generate curvature-dependent motion, in our proposed RD based LSM, the LSE is driven by the reaction function, i.e., the LSE equation. Therefore, we propose to use the diffusion function to regularize the LSF generated by the reaction function. To this end, we propose the following TSSM to solve the RD.

**Step 1:** *Solve the reaction term $\phi_t = -\varepsilon^{-1} L(\phi)$ with $\phi(x,t=0) = \phi^n$ till some time $T_r$ to obtain the intermediate solution, denoted by $\phi^{n+1/2} = \phi(x, T_r)$;*

**Step 2:** *Solve the diffusion term $\phi_t = \varepsilon \Delta \phi$, $\phi(x,t=0) = \phi^{n+1/2}$ till some time $T_d$, and then the final level set is $\phi^{n+1} = \phi(x, T_d)$.*

Although the second step may have the risk of moving the zero level set away from its original position, by choosing a small enough $T_d$ compared to the spatial resolution (i.e., the number of grid points [17]), the zero level set will not be moved [11][17].

In **Step 1** and **Step 2**, by choosing small $T_r$ and $T_d$, we can discretely approximate $\phi^{n+1/2}$ and $\phi^{n+1}$ as $\phi^{n+1/2} = \phi^n + \Delta t_1(-\varepsilon^{-1} L(\phi^n))$ and $\phi^{n+1} = \phi^{n+1/2} + \Delta t_2(\varepsilon \Delta \phi^{n+1/2}))$, respectively, where the time steps $\Delta t_1$ and $\Delta t_2$ represent the time $T_r$ and $T_d$, respectively. Obviously, we can integrate the parameter $\varepsilon$ into the time steps $\Delta t_1$ and $\Delta t_2$ as $\Delta t_1 \leftarrow \Delta t_1(-\varepsilon^{-1})$ and $\Delta t_2 \leftarrow \Delta t_2\varepsilon$, and hence, similar to the diffusion-generated or convolution-generated curvature motion [11][49][51], we only need to consider the two time steps $\Delta t_1$ and $\Delta t_2$ to keep numerical stability. The algorithm of RD based LSE is summarized in **Algorithm 1**.

**4.2 Numerical Implementation**

*A. Numerical approximation for the spatial and time derivatives:* In implementing the traditional LSMs [4][5][10], the upwind scheme is often used to keep numerical stability. By introducing the diffusion term, in the proposed RD-LSE the simple central difference scheme [24] can be used to compute all the spatial partial derivatives $\partial(\cdot)/\partial x_i$, $i = 1,\ldots,n$, and the simple forward difference scheme can be used to compute the temporal partial derivative $\phi_t$.

*B. Setting for the time steps $\Delta t_1$ and $\Delta t_2$:* Since Eq. (19) is a linear PDE, the standard Von Neumann analysis [19][23][24] can be used to analyze the stability for the time step $\Delta t_2$. Putting $\phi_{i,j}^n = r^n e^{I(i\xi_1 + j\xi_2)}$ into Eq. (19), where $I = \sqrt{-1}$ denotes the imaginary unit, we get the amplification factor as

$$r = 1 + 2\Delta t_2 \cdot [\cos(\xi_1) + \cos(\xi_2) - 2] \qquad (20)$$

Therefore, we have $1 - 8\Delta t_2 \leq r \leq 1$. By solving the inequality $|1 - 8\Delta t_2| \leq 1$, we obtain

$$0 \leq \Delta t_2 \leq 0.25 \qquad (21)$$

However, this is a relaxed constraint for $\Delta t_2$ because $\Delta t_2$ also controls the smoothness of LSF during the LSE in **Step 1**, and a large $\Delta t_2$ has the risk of moving the zero level set away from its original position. Therefore, we should use a small enough $\Delta t_2$ compared to the spatial resolution (i.e., the number of grid points [17]) so that the zero level set will not move [11][17][51], and only the LSE force in $L(\phi^n)$ drives the zero level set to evolve (please refer to Fig. 3(b)).

For the time step $\Delta t_1$, since $L(\phi^n)$ in Eq. (18) may contain nonlinear terms, we cannot apply the Von Neumann analysis for stability analysis. However, since the diffusion process in **Step 2** can make the LSF smooth while reducing to some extent the numerical error generated in **Step 1**, we can easily choose a proper $\Delta t_1$ to make the evolution stable. In all our experiments, we set $\Delta t_1 = 0.1$ and it works very well.

*C. Discussion on the evolution speed:* For the re-initialization methods [7-8][10][14], Eq. (5) should be iterated several times to make the LSF be an SDF while keeping the zero level set stationary. This is very time-consuming [9]. The GDRLSE methods are computationally much more efficient than re-initialization method. Refer to Eq. (10), in each iteration the computation of GDRLSE includes two components: the regularization term and LSE term driven by force *F*. In each iteration of our RD method, the computation also includes two similar components. The only difference is that we split the computation into two steps: first compute the LSE term, and then compute the diffusion term. Therefore, the computation complexity of RD is similar to that of GDRLSE methods.

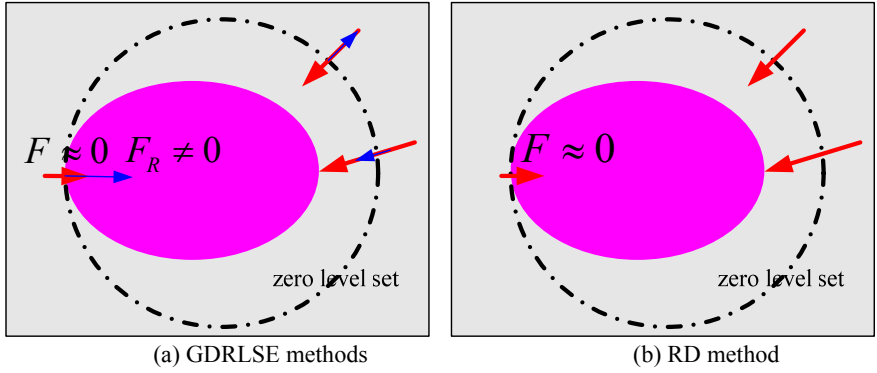

**Fig. 3: LSE force analysis for RD and GDRLSE methods. (a) The possible forces at different positions for GDLRSE methods. The red arrows represent the LSE force *F*, while the blue arrows represent the regularization force $F_R$. (b) In the RD method, only the LSE force *F* (denoted by red arrows) drives the zero level set evolve because we set the time step for the diffusion term small enough to prevent the zero level set moving.**

However, the zero LSE speeds of RD and GDRLSE methods are different. $\delta_{2,\rho}(\phi)$ and $|\nabla\phi|$ are not zero in the neighborhood of zero level set. For GDRLSE methods, when using $\delta(\phi)=\delta_{2,\rho}(\phi)$ we can rewrite Eq. (10) as $\phi_t = F_R \delta_{2,\rho}(\phi) + F\delta_{2,\rho}(\phi)$, where $F_R = \alpha\nabla(r(\phi)\nabla\phi)/\delta_{2,\rho}(\phi)$ is the regularization force that drives the zero level set to evolve. When apply GDRLSE to PDE-based LSM, we can use $|\nabla\phi|$ instead of $\delta(\phi)$, and the similar analysis holds. The zero LSE speed is determined by the total force (i.e., $F_R+F$). However, the signs of diffusion ratios of GDRLSE1 and GDRLSE2 (refer to $r_1$ and $r_2$ in the left figure of Fig. 1) may be inverse for different LSFs during LSE, and this can make the sign of regularization force $F_R$ *different* from that of the LSE force *F* (refer to Fig. 3 (a)), and consequently reduce the evolution speed. When the zero level set reaches the object boundary, the LSE force *F* will be close to zero, and this can make the zero level set finally stop at the object boundary if $F_R=0$. However, when force *F* is zero, the regularization force $F_R$ may not be zero, making the zero level set continue to evolve and finally causing the boundary leakage problem. Such disadvantages also exist for GDRLSE3.

In summary, we cannot set a large time step for GDRLSE methods with $\delta_{2,\rho}(\phi)$ or $|\nabla\phi|$ in order to avoid boundary leakage because the time step affects both regularization force and LSE force. In all our experiments in Section 5, we set time step $\Delta t$=0.1 for GDRLSE methods to alleviate the boundary leakage. We set $\Delta t_1$=0.1 for the first step of our RD method, and set $\Delta t_2$=0.1 or 0.001 for the second step of RD according to the noise level in the image. The small time steps can prevent the zero level set from moving [11][17], and the zero LSE speed is mainly determined by the force $F$ in Eq. (18), making our RD method have better boundary anti-leakage performance than GDRLSE methods (refer to Fig. 3(b)). Overall, sometimes the zero LSE speed of RD is faster than GDRLSE because the LSE force of GDRLSE can be decreased by the regularization force; sometimes the reverse is true because the LSE force of GDRLSE can be increased by the regularization force.

Note that in [9][59], a large time step (e.g., $\Delta t$=5) is set for GDRLSE1 and GDRLSE2 methods, and the LSE is very fast without obvious boundary leakage. However, the good anti-leakage capability of GDRLSE1 and GDRLSE2 comes from not only their regularization term, but also the use of Dirac functional $\delta_{1,\rho}$ that restricts the force function acting only in a small neighborhood of the zero level set (see the red dotted line in the right figure of Fig. 1). Our experimental results also validate this (refer to Figs. 7-9 in Sections 5.3 and 5.4). However, based on our experiments using the codes downloaded from [29] with default settings, it is found that there still exists the risk of boundary leakage even the Dirac functional $\delta_{1,\rho}$ is used in GDRLSE1 (e.g., the middle left image in Fig. 2) and GDRLSE2 if a large time step such as $\Delta t$=5 is set. Some examples are given at http://www.comp.polyu.edu.hk/~cslzhang/RD/RD.htm. Therefore, we set a small time step $\Delta t$=0.1 in all our experiments to alleviate the boundary leakage of GDRLSE1 and GDRLSE2.

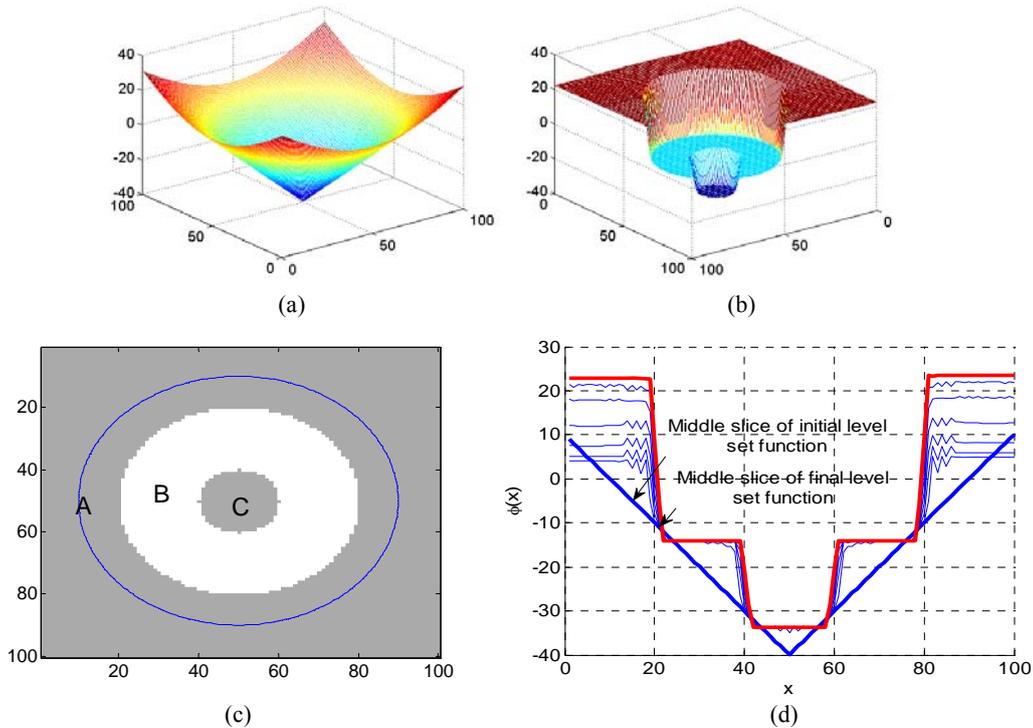

**Fig. 4: The GAC model implemented by the proposed RD method on an image with interior boundary. (a) Initial level set function. (b) Final level set function. (c) Testing image. Blue circle represents the initial contour. There are three regions: A, B and C. (d) Middle slices of level set function during LSE. The red solid line represents the middle slice of the final level set function, which is piecewise constant in each region (A, B or C). We set $\Delta t_1$=0.1 and $\Delta t_2$=0.001.**

### 4.3 The Consistency between Theory and Implementation

For images contaminated by strong noise, the diffusion rate should be set a little large to remove noise, and hence the final LSF tends to be piecewise smooth (see the bottom left second figure in Fig. 10 for an example). This does not contradict **Theorem 2**, which indicates that the final LSF tends to be piecewise constant (stronger than piecewise smooth), because the diffusion rate is assumed very small in **Theorem 2**. On the other hand, as long as the time step $\Delta t_2$ is chosen small enough compared to the spatial resolution, it can be guaranteed that the zero level set will not move [17]. Since the interface evolution is independent of the initial functions provided that their zero level sets are at the same position [8], the zero level set of the final steady state solution by the proposed RD method will keep unchanged.

As an example, we use the proposed RD method to implement the GAC model, and then apply it to an artificial image with interior boundary. The results are shown in Fig. 4. We set $\Delta t_2$=0.001, which is small enough to ensure the final LSF to be a piecewise constant function. It can be seen from Fig. 4 (d) that the LSF is gradually tending to be piecewise constant so that the value in each region is nearly a constant during LSE, which is consistent with **Theorem 2**.

For region-based models such as the CV model [18], since the data terms often take large values which weaken the effect of diffusion term, we can set a large diffusion rate to keep the LSF smooth during LSE, and hence the final LSF is often piecewise smooth (see Figs. 10 for examples).

### 5. EXPERIMENTAL RESULTS

#### 5.1. Setup of Experiments

In our experiments, all the competing methods use the same level set model, while the only differences are the different regularization terms used in them. As explained in Section 2.2 B, we use GDRLSE1, GDRLSE2 and GDRLSE3 to represent the methods in [9], [59] and [34], respectively. The level set models used in our experiments are summarized in Table I. In the following experiments, in most cases we initialize the LSF to be a binary function whose values have positive and negative signs respectively inside and outside the contour. We set $\rho = 0.5$ for GDRLSE3 as suggested by [34] and $\rho =1$ for other methods. Other parameters are set according to the different experiments. The Matlab source code of the proposed RD method and more experimental results can be found in http://www.comp.polyu.edu.hk/~cslzhang/RD/RD.htm.

TABLE I: THE MODELS USED IN EXPERIMENTS. $c$ IS A CONSTANT, $\kappa$ IS THE CURVATURE OF LSF, $g$ IS AN EDGE INDICATOR FUNCTION [5], AND $\lambda, \nu, \alpha, \mu, \lambda_1, \lambda_2 > 0$ ARE FIXED PARAMETERS.

| Section | Force term $F$ | GDRLSE1 | GDRLSE2 | GDRLSE3 | RD |
|---|---|---|---|---|---|
| 5.2 | $F = c$, or $\kappa$ | -- | -- | -- | $\phi_t = F\|\nabla\phi\|$ |
| 5.3 | $F = \lambda \mathrm{div}\left(g\dfrac{\nabla\phi}{\|\nabla\phi\|}\right) + \nu g$ | $\phi_t = \alpha\mathrm{div}[r_1(\phi)\nabla\phi] + F\delta_{2,\rho}(\phi)$ (Fig. 7) <br> $\phi_t = \alpha\mathrm{div}[r_1(\phi)\nabla\phi] + F\delta_{1,\rho}(\phi)$ (Fig. 8) | $\phi_t = \alpha\mathrm{div}[r_2(\phi)\nabla\phi] + F\delta_{2,\rho}(\phi)$ (Fig. 7) <br> $\phi_t = \alpha\mathrm{div}[r_2(\phi)\nabla\phi] + F\delta_{1,\rho}(\phi)$ (Fig. 8) | $\phi_t = \alpha\mathrm{div}[r_3(\phi)\nabla\phi] + F\delta_{2,\rho}(\phi)$ (Fig. 7) <br> $\phi_t = \alpha\mathrm{div}[r_3(\phi)\nabla\phi] + F\delta_{1,\rho}(\phi)$ (Fig. 8) | $\phi_t = F\delta_{2,\rho}(\phi)$ (Fig. 7) <br> $\phi_t = F\delta_{1,\rho}(\phi)$ (Fig. 8) |
| 5.4 | $F = \mathrm{div}\left(g\dfrac{\nabla\phi}{\|\nabla\phi\|}\right) + \nu g$ | $\phi_t = \alpha\mathrm{div}[r_1(\phi)\nabla\phi] + F\|\nabla\phi\|$ | $\phi_t = \alpha\mathrm{div}[r_2(\phi)\nabla\phi] + F\|\nabla\phi\|$ | $\phi_t = \alpha\mathrm{div}[r_3(\phi)\nabla\phi] + F\|\nabla\phi\|$ | $\phi_t = F\|\nabla\phi\|$ |
| 5.5 | $F = \mu\mathrm{div}\left(\dfrac{\nabla\phi}{\|\nabla\phi\|}\right) - \nu - \lambda_1(I - c_{\mathrm{in}}(\phi))^2 + \lambda_2(I - c_{\mathrm{out}}(\phi))^2$ | $\phi_t = \alpha\mathrm{div}[r_1(\phi)\nabla\phi] + F\delta_{2,\rho}(\phi)$ | $\phi_t = \alpha\mathrm{div}[r_2(\phi)\nabla\phi] + F\delta_{2,\rho}(\phi)$ | $\phi_t = \alpha\mathrm{div}[r_3(\phi)\nabla\phi] + F\delta_{2,\rho}(\phi)$ | $\phi_t = F\delta_{2,\rho}(\phi)$ |
| 5.6 | The models are those used in Section 5.3, Section 5.4 and Section 5.5, respectively. | | | | |

We first apply the RD method to PDE-based LSM to demonstrate its superior performance to re-initialization methods; second, we apply it to edge-based variational level set models with different Dirac functionals and compare it with GDRLSE methods for images with weak boundaries; third, we apply the RD method to classical GAC model [5] and the CV model [18] in comparison with GDRLSE and representative LSMs with re-initialization; finally, we quantitatively compare RD with GDRLSE methods for the edge-based variational level set model, the PDE-based GAC model and the region-based CV model. The advantages of our RD method over re-initialization methods and GDRLSE methods are summarized as follows.

A. The RD method can keep the LSE process stable for both variational LSM and PDE-based LSM, and it is much more efficient than re-initialization method (refer to Figs. 5 and 6 in Section 5.2).

B. The *edge-based* LSMs mainly have three types: the variational LSM with $\delta_{2,\rho}$ or $\delta_{1,\rho}$, and the PDE-based LSM such as GAC model. For variational LSM with $\delta_{2,\rho}$, in general our RD method has better boundary anti-leakage ability than GDRLSE methods (refer to Fig. 7 in Section 5.3). When Dirac functional $\delta_{1,\rho}$ is used for the edge-based variational LSM, RD, GDRLSE1 and GDRLSE2 all have good boundary anti-leakage performance (refer to Fig. 8 in Section 5.3). For the PDE-based GAC model, the boundary anti-leakage performance of our RD method is much better than re-initialization method and GDRLSE methods (refer to Fig. 9 in Section 5.4). This is because in RD-LSE only the LSE force $F$ drives the zero level set to evolve (see Fig. 3(b)), while for GDRLSE methods the regularization force $F_R$ may be nonzero at the object boundary and drive the zero level set passing the object boundary (see Fig. 3(a)).

*C.* For *region-based* LSM such as the CV model, our RD method has better anti-noise performance than the re-initialization method and GDRLSE methods (refer to Fig. 10 in Section 5.5).

*D.* From the quantitative evaluation results in Section 5.6, one can see that our RD method has much better overall performance than the re-initialization method and GDRLSE methods (refer to Figs. 12 and 13 in Section 5.6).

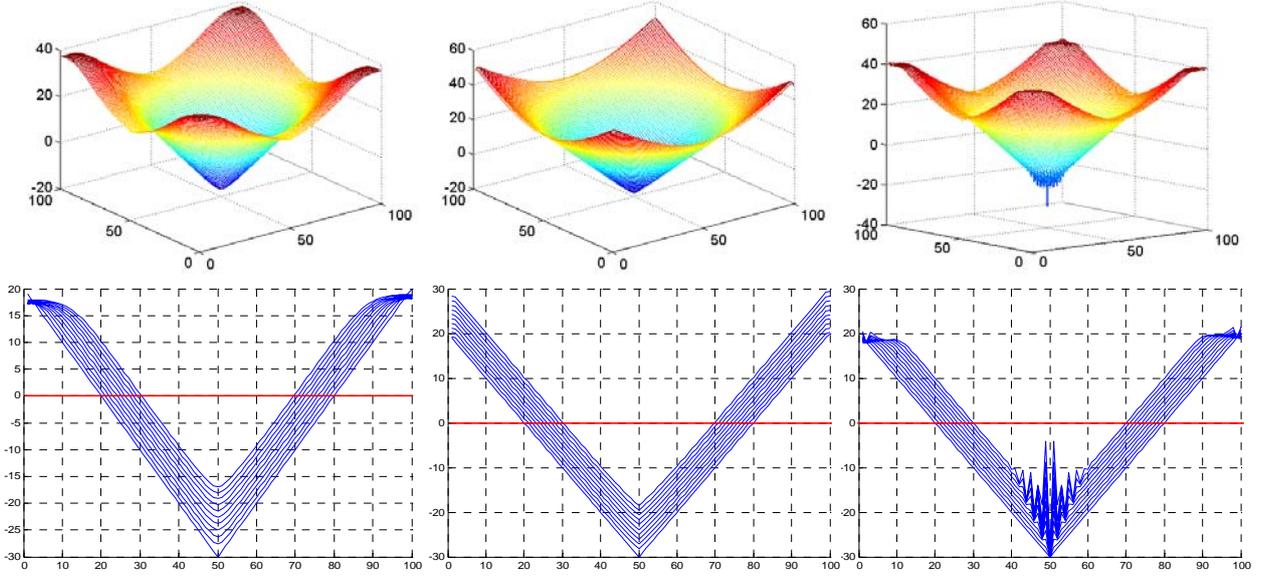

**Fig. 5: Top row: the final LSFs. Bottom row: the middle slices of the LSFs in iterations. From left to right: results by RD method, re-initialization method and the direct implementation without re-initialization. We set $\Delta t_1 = \Delta t_2 = 0.1$.**

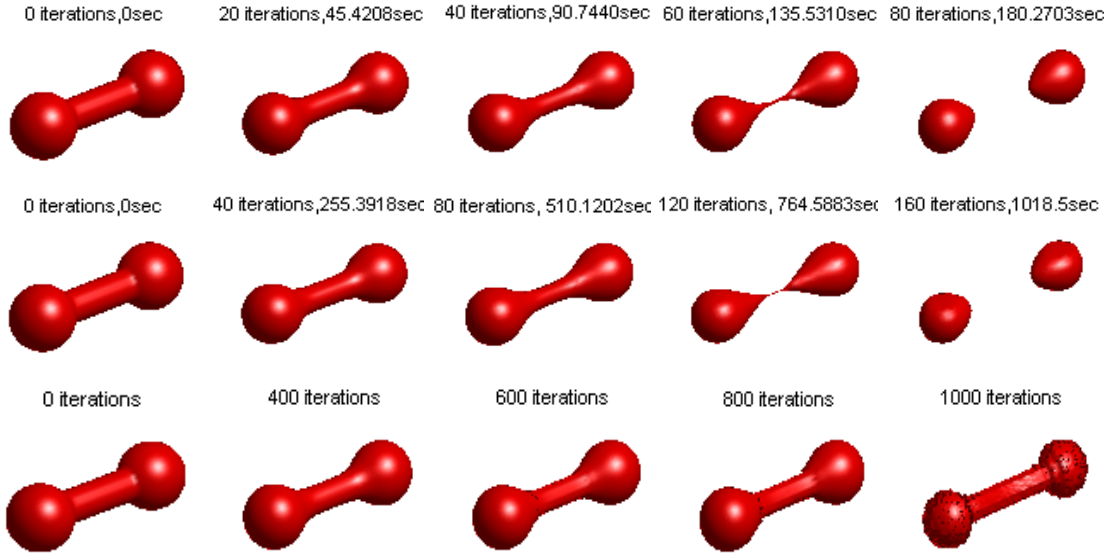

**Fig. 6: Motion of dumbbell driven by mean curvature. Top row: LSE process with RD ($\Delta t_1 = \Delta t_2 = 0.1$); Middle row: LSE process with re-initialization; Bottom row: LSE process without re-initialization ($\Delta t_1 = 0.1$).**

## 5.2 Experiments on PDE-based Level Set Method

We consider a simple case of shrinking a circle with force function $F=1$ according to Eq. (3). The initial LSF is $\phi_0(x_1, x_2) = \sqrt{(x_1-50)^2 + (x_2-50)^2} - 30$. We set the time steps as $\Delta t_1 = \Delta t_2 = 0.1$, and the spatial steps are $\Delta x_1 = \Delta x_2 = 1$. The number of iterations is 100.

The middle slices of the LSFs during LSE and the final LSFs are shown in Fig. 5. We see that when the zero level set moves to the center, the direct implementation without re-initialization [8] leads to serious spikes, making the computation highly inaccurate. In contrast, the proposed RD method does not have such a problem, and the LSF can always keep smooth during evolution, ensuring an accurate computation. The bottom left figure of Fig. 5 also demonstrates that RD will not move the zero level set by using a small diffusion rate. The bottom middle figure in Fig. 5 shows the results by traditional LSMs with re-initialization [5][7][8][10]. It is obvious that both re-initialization and RD can ensure the evolution stable. However, RD has much less computational cost and it is much easier to implement.

In Fig. 6, we test the RD method on 3D LSE driven by mean curvature. The LSE equation is as that in Eq. (3), where $F=\kappa=\mathrm{div}(\nabla\phi/|\nabla\phi|)$ and $\phi(x): R^3 \rightarrow R$. Consider the initial LSF as the shape of a dumbbell with two large spheres connected by a cylinder. The RD method can keep the LSE stable, while for the direct implementation without re-initialization, the LSF becomes unsmooth during LSE and the evolution becomes unstable. For the re-initialization method, although the evolution can keep stable, it is much more time-consuming than our RD approach in terms of both the number of iterations and time.

## 5.3 Experiments with Edge-based Variational LSM

In [9][59], the Dirac functional is approximated by $\delta_{1,\rho}$ and the force term is $F=\lambda\mathrm{div}(g(|\nabla I|)\nabla\phi/|\nabla\phi|)+\nu g(|\nabla I|)$, where $\lambda$ and $\nu$ are fixed parameters and $g(|\nabla I|)$ is an edge indicator function. In Fig. 2, $\delta_{1,\rho}$ is used, and GDRLSE1, GDRLSE2 and the proposed RD lead to similar results. As explained in Section 2.2-C, to validate more comprehensively the performance of a method, other Dirac functional should also be considered. In this sub-section, we first approximate the Dirac functional by $\delta_{2,\rho}$ and compare RD with GDRLSE methods on images with weak boundaries, and then we make more tests by using Dirac functional $\delta_{1,\rho}$. Our experiments validate that the boundary anti-leakage performance of GDRLSE methods is much affected by the Dirac functional.

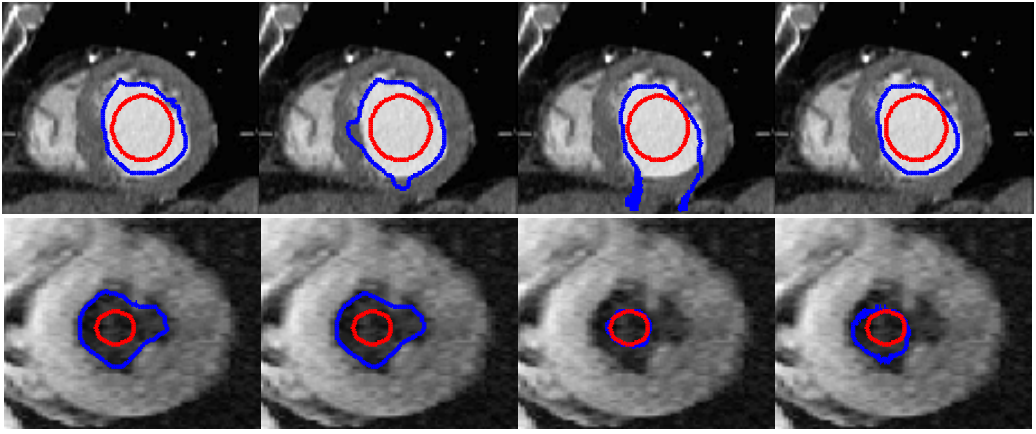

**Fig. 7: Segmentation results on two left ventricle images by using edge-based variational LSM with $\delta_{2,\rho}$. From left to right: results by RD, GDRLSE1, GDRLSE2 and GDRLSE3. The red circles represent the initial contours, and the blue solid curves represent the final contours. The parameters are set as $\Delta t_1$=0.1, $\Delta t_2$=0.001, $\alpha$=0.2, $\lambda$=1, $\nu$=3.5.**

Fig. 7 shows some segmentation results by RD and GDRLSE methods for edge-based variational LSM with $\delta_{2,\rho}$. The test images are two magnetic resonance images of the left ventricle of a human heart. For the first image, both RD and GDRLSE3 yield satisfying segmentation results without much boundary leakage, but GDRLSE1 and GDRLSE2 lead to serious boundary leakage because the boundary of this image is very weak and blurred. For the other image, both RD and GDRLSE1 generate satisfying results. However, for GDRLSE2 and GDRLSE3, the LSE falls into local minima because the image data force terms are significantly affected by the noisy objects.

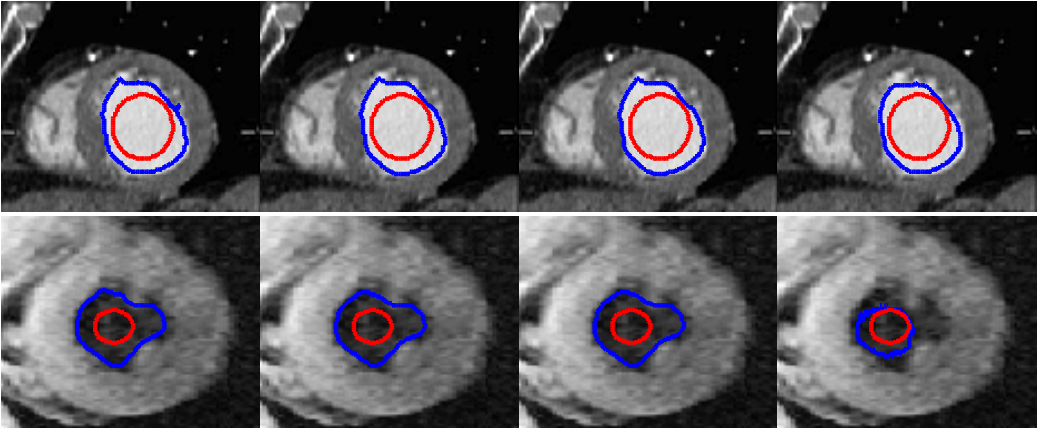

**Fig. 8: Segmentation results on two left ventricle images by using edge-based models with $\delta_{1,\rho}$. From left to right: results by RD, GDRLSE1, GDRLSE2 and GDRLSE3. The red circles represent the initial contours, and the blue solid curves represent the final contours. The parameters are set as $\Delta t_1$=0.1, $\Delta t_2$=0.001, $\alpha$=0.2, $\lambda$=1, $\nu$=3.5.**

Fig. 8 shows the segmentation results of the two images by using $\delta_{1,\rho}$. We can see that RD, GDRLSE1, and GDRLSE2 all yield satisfying segmentation results. The segmentation results by GDRLSE1 and GDRLSE2 are much better than their results with $\delta_{2,\rho}$ in Fig. 7. This validates that $\delta_{1,\rho}$ has a better boundary anti-leakage capability than $\delta_{2,\rho}$. However, $\delta_{1,\rho}$ limits the LSF to evolve only around the zero level set and thus prevents the emerging of new contours, making the LSE easily fall into local minima. More experimental results by using edge-based models with $\delta_{1,\rho}$ and $\delta_{2,\rho}$ can be found at the website associated with this paper: http://www4.comp.polyu.edu.hk/~cslzhang/RD/RD.htm.

**5.4 Experiments with the PDE-based GAC Model**

In the LSE equation (refer to Eq. (3)) of GAC model [5], there is $F=\text{div}(g(|\nabla I|)\nabla\phi/|\nabla\phi|)+\nu g(|\nabla I|)$, where $g(|\nabla I|)$ is an edge indicator function and $\nu$ is a fixed parameter. The LSE equation of GAC model is PDE-based. We adapted the GDRLSE methods to the GAC model and compared them with our RD method.

Fig. 9 shows the segmentation results on a noisy synthetic image with weak boundaries. The re-initialization method [8][10] can keep the LSF smooth during LSE, and hence reduce the numerical error to some extent. However, the re-initialization will move the zero level set away from its original position, resulting in boundary leakage [38]. (In order to be consistent with the traditional LSMs [2-5][7][10][18], we initialized the LSF to be an SDF for the re-initialization method.) As shown in Fig. 9, the RD method results in a very good segmentation without boundary leakage, and the final LSF can be approximated as a piecewise constant function with constant values inside and outside the contours. This is again consistent with what is claimed in **Theorem 2**. For GDRLSE1, some contours occur inside the object during evolution because $|\nabla\phi|$ acts on all level curves, making the LSF evolve in the whole domain and produce false peaks and valleys. For GDRLSE2, obvious boundary leakage occurs. This is also because in the LSE equation of GAC model in Eq. (10), $\delta(\phi)$ is replaced by $|\nabla\phi|$ which acts on all level curves, leading to boundary leakage. Similarly, boundary leakage occurs for GDRLSE3.

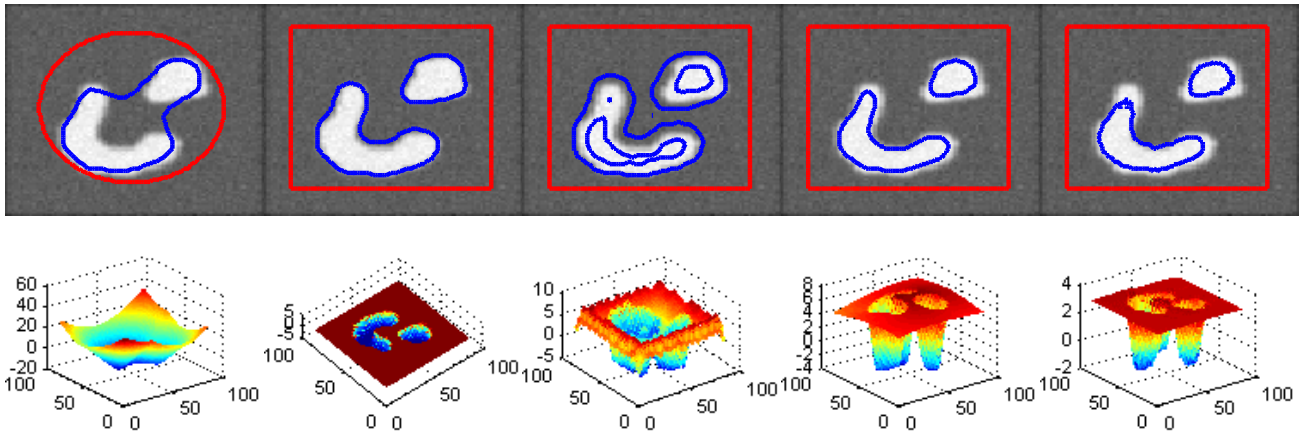

**Fig. 9: Segmentation results on a synthetic image (downloaded from [29]). From left to right: results by re-initialization method [8], RD, GDRLSE1, GDRLSE2 and GDRLSE3. The red curves represent the initial contours, and the blue solid curves represent the final contours. We set parameters $\Delta t_1$=0.1, $\Delta t_2$=0.001, $\nu$=0.5, $\alpha$=0.2.**

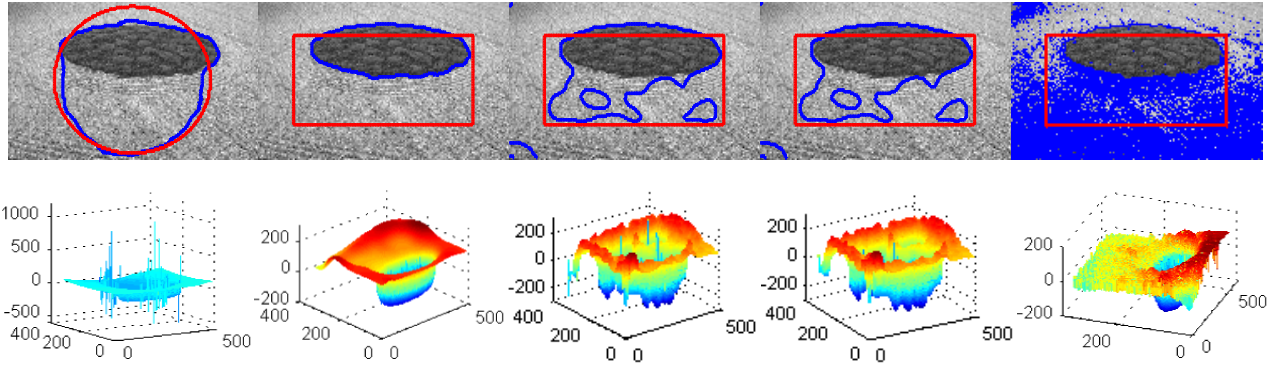

**Fig. 10**: Segmentation results on a real image with noisy background (downloaded from [46]). From left to right: results by re-initialization method [8], RD, GDRLSE1, GDRLSE2 and GDRLSE3. Top row: initial contours (red curves) and final contours (blue curves). Bottom row: final LSFs. We set the same parameters $\Delta t_1 = 0.1$, $\mu=0.5\times 255^2$, $\nu=0$, $\lambda_1=\lambda_2=1$ for all the methods, and set parameter $\Delta t_2=0.1$ for our RD method. For GDRLSE methods, we set parameter $\alpha=0.2$.

We also tested some images from the Coral dataset [46], and the results can be found at the website associated with this paper.

### 5.5 Experiments with the CV Model

The CV model [18] is a simplified Mumford-Shah model [43], which assumes that the image is piecewise constant. The LSE equation of the CV model is in Eq. (4), where $F=\mu \text{div}(\nabla\phi/|\nabla\phi|)-\nu-\lambda_1(I-c_{\text{in}}(\phi))^2 + \lambda_2(I-c_{\text{out}}(\phi))^2$, $\mu,\nu\geq 0$; $\lambda_1,\lambda_2>0$ are fixed parameters, $c_{\text{in}}(\phi)$ and $c_{\text{out}}(\phi)$ are average intensity inside and outside zero level set, respectively. The Dirac functional is approximated by $\delta_{2,\rho}$ defined in Eq. (12). In implementing the LSE equation, the re-initialization is optional [18]. However, without re-initialization the LSF will become unsmooth during LSE, especially when the image is noisy, and hence serious numerical error can occur. Therefore, in the following experiments, we mainly compare the RD method with the re-initialization method [8] and the three GDRLSE methods.

*A. Experiments on noisy images:* Fig. 10 shows the segmentation results on a real image with noisy background. The noisy background results in big numerical errors for methods without re-initialization, making the LSF fail to evolve stably. For the re-initialization method [8], it fails to re-initialize the LSF to be an SDF because of the strong noise. As shown in the left second column of Fig. 10, the RD method can still yield desirable results, and the final LSF keeps smooth. This is because the diffusion procedure in the second step of our RD algorithm can reduce the error produced by the LSE in the first step effectively. Since we used a large diffusion rate $\Delta t_2=0.1$, the final LSF is nearly piecewise smooth. The GDRLSE1 method falls into local minima because the noisy background makes the regularization term invalid, and there exist some spikes in the final LSF. The GDRLSE2 method yields similar results to GDRLSE1, because the large gradient of the LSF caused by the noisy force term makes the two different regularization terms in GDRLSE1 and GDRLSE2 perform nearly the same (please refer to the definitions of $r_1$ and $r_2$ in Eq. (7) and Eq. (8), respectively, and we can see that when $|\nabla \phi|$ is large, $r_1$ and $r_2$ are the same). For GDRLSE3, the final segmentation result is very noisy because the diffusion rate $r_3(\phi)=\mathcal{H}_\rho(|\nabla\phi|-1)$ changes smoothly from 0 to 1, which limits its regularization capability. More experimental results on some real images can be found in the website of this paper.

***B. Demonstration of global minimum:*** To further demonstrate that our RD method with the CV model can reach the global minimum while being robust to level set initializations, we apply it to a noisy synthetic image with different level set initializations, as shown in Fig. 11. The zero contours can be set outside the objects, around all the objects, cross some objects or even inside one object. Although these initializations are very different, the final contours are almost the same, which validates that our RD method can robustly evolve to the global minimum of the energy functional, leading a good global segmentation.

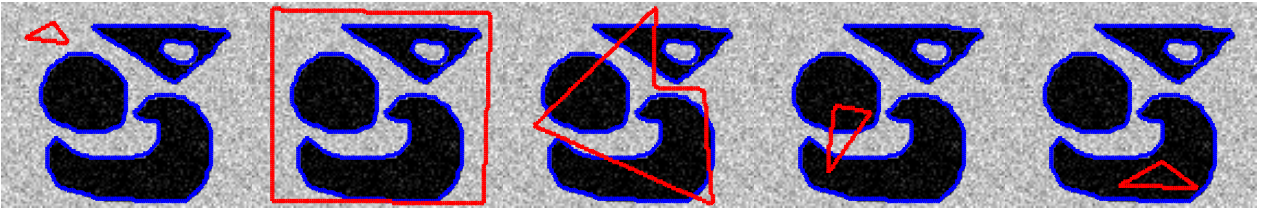

**Fig. 11: Segmentation results on a noisy synthetic image with different level set initializations. Red curves denote the initial contours and blue curves denote the final contours. We set the parameters $\Delta t_1$ =0.1, $\Delta t_2$ =0.1, $\mu$=0.001×255$^2$, $\nu$=0, $\lambda_1$=$\lambda_2$=1 for all images.**

**5.6 Quantitative Experiments**

We use the Jaccard similarity (JS) [63] as an index to evaluate quantitatively the segmentation performance of our RD method, re-initialization method and GDRLSE methods. The *JS* between two regions $S_1$ and $S_2$ is calculated by $J(S_1, S_2) = \frac{|S_1 \cap S_2|}{|S_1 \cup S_2|}$, which is the ratio between the intersectional area of $S_1$ and $S_2$ and their united area. Obviously, the closer the *JS* value is to 1, the more similar $S_1$ is to $S_2$. In our experiments, $S_1$ is the segmented region by the five competing methods, and $S_2$ is the ground truth. Due to the randomness of added noise, we run the program 50 times, and then calculated the average of the *JS* values.

Fig. 12 shows the *JS* indices by applying the five competing methods to a synthetic image and its noisy versions. The used level set models are edge-based variational model with different Dirac functionals (refer to Section 5.3) and the GAC model (refer to Section 5.4). We use the same binary function to initialize the LSF for all the methods except for the re-initialization method, which uses an SDF. The purpose of this experiment is to test the anti-noise performance of the five methods. From Fig. 12, we can see that the performance of the re-initialization and GDRLSE methods fluctuates significantly for the noise with different strength. However, the *JS* value by our RD method does not change much, which demonstrates the robust anti-noise performance of RD. We can also see that the performance of GDRLSE1 and GDRLSE2 with $\delta_{1,\rho}$ is much better than those with $\delta_{2,\rho}$, which again validates that GDRLSE1 and GDRLSE2 can be severely affected by using a different Dirac functional (e.g., $\delta_{2,\rho}$) from the one (i.e., $\delta_{1,\rho}$) used in [59]. Nonetheless, with $\delta_{1,\rho}$ GDRLSE1 and GDRLSE2 still have a risk of boundary leakage, as discussed in Section 4.2-C.

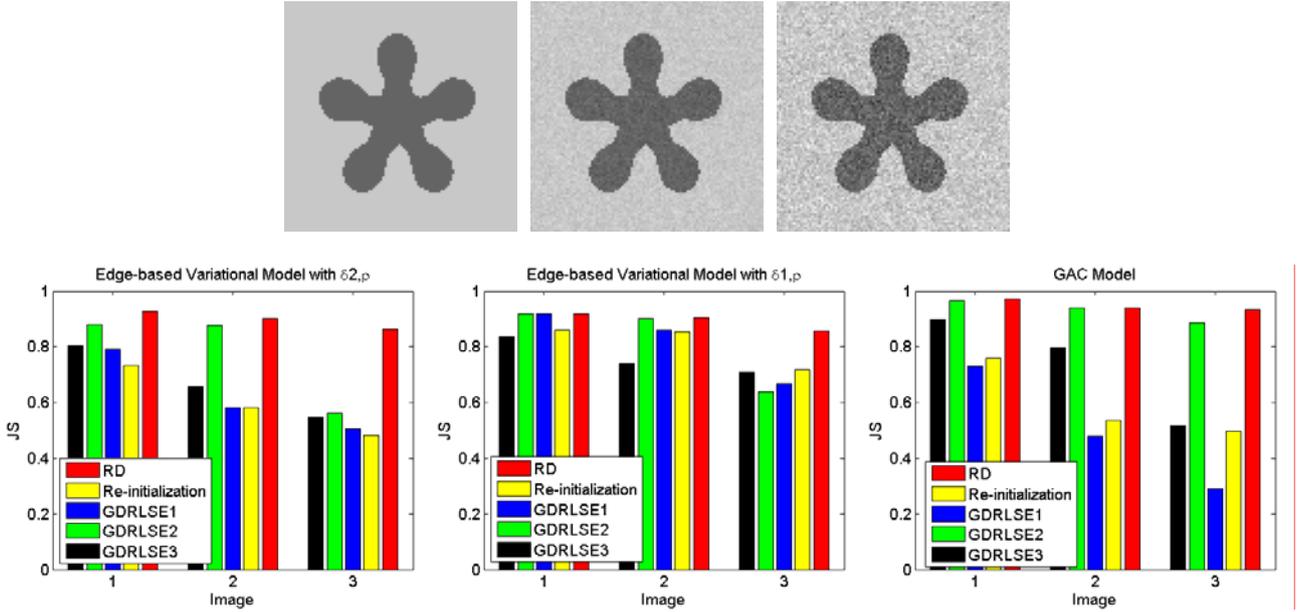

**Fig. 12: Quantitative comparisons among RD and GDRLSE methods for edge-based models. Top row, from left to right: clean image (image 1), noisy image (image 2) (Gaussian noise with zero mean and standard deviation $\sigma=0.001$), and noisy image (image 3) (Gaussian noise with zero mean and standard deviation $\sigma=0.005$). Bottom row, from left to right: the *JS* values using the edge-based variational model with $\delta_{2,\rho}$ and $\delta_{1,\rho}$ in Section 5.3, and GAC model in Section 5.4, respectively. For edge-based variational models, we set $\Delta t_1=0.1$, $\Delta t_2=0.001$, $\alpha=0.2$, $\lambda=1$, $\nu=0.05$ for all the three images. For the GAC model, we set $\Delta t_1=0.1$, $\Delta t_2=0.001$, $\alpha=0.2$, $\lambda=1$ for all the three images, and we set $\nu=0.05$, $\nu=0.2$, $\nu=0.5$ for the images from left to right.**

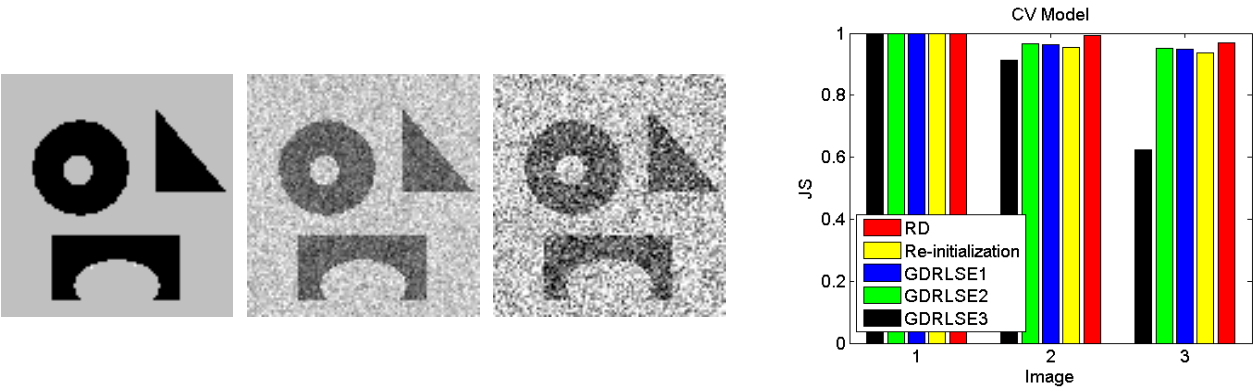

**Fig. 13: Quantitative comparisons among RD, re-initialization, and GDRLSE methods for the CV model [18]. Left three images: clean image, images with Gaussian noise of zero mean and standard deviation $\sigma=0.01$, $\sigma=0.05$, respectively. Right image: the *JS* values by competing methods. We set $\Delta t_1=0.1$, $\Delta t_2=0.01$, $\mu=0.1\times 255^2$, $\nu=0$, $\lambda_1=\lambda_2=1$ for all three images.**

Fig. 13 shows the quantitative comparison results by applying the five methods to the CV model for segmenting a synthetic image with different noise levels. We can see that when the noise is not very strong, all methods have satisfying results. When strong noise is added, the performance of GDRLSE3 degenerates dramatically; the results by GDRLSE1 and GDRLSE2 are almost the same, and this shows that the regularization capabilities of the two methods are similar when applied to the CV model. The *JS* index by RD is the highest among all the five methods, demonstrating the superior anti-noise performance of RD to other methods. Moreover, it can also be seen that the segmentation results by the CV model depend on not only the region-based data force term but also the different regularization terms.

**TABLE II: Iterations (Iter) and CPU time (in seconds) by RD and GDRLSE methods. The values in bold represent the best results.**

| Methods | Fig.12 Image 1 | | Fig.12 Image 2 | | Fig.12 Image 3 | | Fig.13 Image 1 | | Fig.13 Image 2 | | Fig.13 Image 3 | |
|---|---|---|---|---|---|---|---|---|---|---|---|---|
| | Size: 100×100 pixels | | | | | | Size: 96×101 pixels | | | | | |
| | Edge-based model with $\delta_{1,\rho}$ | | | | | | CV model | | | | | |
| | Time(s) | Iter | Time(s) | Iter | Time(s) | Iter | Time(s) | Iter | Time(s) | Iter | Time(s) | Iter |
| RD | **4.8** | **500** | **5.0** | **520** | **5.8** | 600 | 4.8 | 150 | **5.1** | **160** | **5.1** | **160** |
| GDRLSE1 | 6.2 | 550 | 6.5 | 580 | 6.7 | **590** | 5.1 | 150 | 9.3 | 200 | 10.9 | 230 |
| GDRLSE2 | 9.3 | 630 | 9.7 | 660 | 10.1 | 690 | 4.7 | **130** | 11.4 | 200 | 12.9 | 230 |
| GDRLSE3 | 14.2 | 1000 | 17.0 | 1200 | 17.8 | 1250 | **4.0** | **130** | 18.1 | 210 | 68.9 | 800 |

At last, let's compare the efficiency of our RD method with other GDRLSE methods in terms of converged iterations and CPU time. All the competing methods are run under Matlab R2010a programming environment in a desktop with Windows XP OS, Pentium Dual-Core 2.10 GHz CPU and 1.95 GB RAM. From our extensive experiments, it can be observed that only the edge-based models with Dirac functional $\delta_{1,\rho}$ can yield good results for most GDRLSE methods. Moreover, by using the CV model the GDRLSE methods can lead to good results. Therefore, we evaluate the efficiency for edge-based models with Dirac functional $\delta_{1,\rho}$ and the CV model. It should be noted that different parameter settings and level set initializations will affect much the efficiency of LSE. For a fair comparison, we choose the synthetic images in Fig. 12 and Fig. 13 for evaluation since all the competing methods can yield favorable results on them. We tune the parameters for each method so that their best results can be obtained. The comparison results are illustrated in Table II. It can be seen that in average the RD method achieves the best performance in terms of both iterations and CPU time. In summary, the proposed RD method has high computational efficiency while having high segmentation accuracy and robustness.

## 6. CONCLUSIONS AND DISCUSSIONS

In this paper, we proposed a reaction-diffusion (RD) based level set evolution (LSE), which is completely free of the re-initialization procedure required by traditional level set methods. A two-step-splitting-method (TSSM) was then proposed to effectively solve the RD based LSE. The proposed RD method can be generally applied to either variational level set methods or PDE-based level set methods. It can be implemented by using the simple finite difference scheme. The RD method has the following advantages over the traditional level set method and state-of-the-art algorithms [9][59][34]. First, the RD method is general, which can be applied to the PDE-based level set methods and variational ones. Second, the RD method has much better performance on weak boundary anti-leakage. Third, the implementation of the RD equation is very simple and it does not need the upwind scheme at all. Fourth, the RD method is robust to noise. The experiments on synthetic and real images demonstrated the promising performance of our approach.

Motivated by the convolution-generated curvature motion [48-51] in phase transitions, the diffusion procedure in our TSSM algorithm can also be replaced by convolving any positive, radically symmetric kernel with a small enough width. Actually we have used a Gaussian kernel to regularize the level set function in our previous work [52]. In our previous work [53], we utilized a constant kernel to regularize two level set functions and achieved promising results. This implies that the RD method can be readily extended to multiphase level set method based on the theory of phase transitions in mixtures of Cahn-Hilliard fluids [26].

# Appendix A. Proof of Theorem 1

**Theorem 1:** *Let $\Omega \subset R^n$, n=2 or 3, is the domain of the level set function $\phi$ and assume that $E(\phi)$ is an energy functional w.r.t. $\phi$, the Euler equations of $E(\phi)$ and $F(\phi)$ are the same, i.e. $E_\phi(\phi)=F_\phi(\phi)$, where $F(\phi) \triangleq \int_\Omega E(\phi) d\mathrm{x}$.* ■

*Proof*: It is very easy to validate this theorem. The energy functional $E(\phi)$ is a constant when $\phi$ is chosen. Therefore, $F(\phi) \triangleq \int_\Omega E(\phi) d\mathrm{x} = E(\phi) S_\Omega$, where $S_\Omega$ is the area of the domain $\Omega$ which is a constant. Thus, it is easy to yield the conclusion $E_\phi(\phi) = F_\phi(\phi)$. ■

# Appendix B. Proof of Theorem 2

**Theorem 2:** *If there are $k \geq 2$ local minima $c_1, \ldots, c_k$ for the energy functional $E(\phi) \geq 0$ in the domain $\Omega$ such that $\{E(c_i)=0, i=1, \ldots, k\}$, then for the point $\mathrm{x}$ where the initial function $\phi_0(\mathrm{x})$ is in the basin of attraction of $c_i$, the solution $\phi(\mathrm{x}, t, \varepsilon)$ of $P_\varepsilon$ will approach to $c_i$ as $\varepsilon \to 0^+$, which is also the equilibrium solution of the LSE equation $\phi_t = -E_\phi(\phi)$ with the same initialization $\phi(\mathrm{x}, t=0) = \phi_0(\mathrm{x})$, i.e.,*

$$\lim_{t \to +\infty, \varepsilon \to 0^+} \phi(\mathrm{x}, t, \varepsilon) = \sum_{i=1}^{k} c_i \chi_i(\mathrm{x})$$

*where $\chi_i(\mathrm{x}) \in \{0,1\}$ is the characteristic function of the set $S_i = \{\mathrm{x} | \phi_0(\mathrm{x}) \in B_i, i=1, \ldots, k\}$, where $B_i$ is a basin to attract $\phi(\mathrm{x}, t, \varepsilon)$ to $c_i$.* ■

*Proof*: The proof of **Theorem 2** is mainly motivated by [20]. The LSE equation for Eq. (16) is

$$\begin{cases} \phi_t = \varepsilon\Delta\phi - \dfrac{1}{\varepsilon}E_\phi(\phi), \text{x} \in \Omega \subset R^n \\ \text{subject to } \phi(\text{x},t=0,\varepsilon) = \phi_0(\text{x}) \end{cases} \quad (22)$$

where $E_\phi(\phi)$ denotes the Gateaux derivative (or first variation) of the energy functional $E(\phi)$ [22]. We assume that $E(\phi)$ and the boundary $\partial\Omega$ are sufficiently smooth. Since we are only interested in the case when $\varepsilon \to 0^+$, by introducing the new time variable $\tau = t/\varepsilon$ and using Taylor's expansion w.r.t. $\varepsilon$, we can write $\phi$ in the form

$$\phi(\text{x},t,\varepsilon) = v(\text{x},\tau,\varepsilon) = v_0(\text{x},\tau) + \varepsilon v_1(\text{x},\tau) + \varepsilon^2 v_2(\text{x},\tau) + O(\varepsilon^3) \quad (23)$$

The derivative of Eq. (23) w.r.t. the time *t* is as follows

$$\phi_t = \frac{1}{\varepsilon}\frac{\partial v}{\partial \tau} = \frac{1}{\varepsilon}\left(\frac{\partial v_0}{\partial \tau} + \varepsilon\frac{\partial v_1}{\partial \tau} + \varepsilon^2\frac{\partial v_2}{\partial \tau} + O(\varepsilon^3)\right) \quad (24)$$

Putting Eq. (24) into Eq. (22), we obtain $\varepsilon\dfrac{\partial v_2}{\partial \tau} + \dfrac{1}{\varepsilon}\dfrac{\partial v_0}{\partial \tau} + \dfrac{\partial v_1}{\partial \tau} + O(\varepsilon^2) = \varepsilon\Delta\phi - \dfrac{1}{\varepsilon}E_\phi(\phi)$. Letting $\varepsilon \to 0^+$ and comparing the coefficients of the power of $\varepsilon$ in both sides, we obtain

$$\frac{\partial v_0}{\partial \tau} = -E_\phi(v_0) \quad (25)$$

Since for $\forall \varepsilon$, $\phi(\text{x},t=0,\varepsilon) = \phi_0(\text{x}) = v_0(\text{x},\tau) + \varepsilon v_1(\text{x},\tau) + \varepsilon^2 v_2(\text{x},\tau) + O(\varepsilon^3)$, comparing the coefficients of the power of $\varepsilon$, we have $v_0(\text{x},\tau=0) = \phi_0(\text{x})$, $v_1(\text{x},\tau=0) = v_2(\text{x},\tau=0) = 0$.

Rearranging Eq. (24), we obtain

$$\frac{\partial v_1}{\partial \tau} = \phi_t - \frac{1}{\varepsilon}\frac{\partial v_0}{\partial \tau} - O(\varepsilon) = \phi_t + \frac{1}{\varepsilon}E_\phi(v_0) - O(\varepsilon) \quad (26)$$

Putting Eqs.(22) and (25) into Eq. (26), we obtain

$$\begin{cases} \dfrac{\partial v_1}{\partial \tau} = \varepsilon\Delta\phi - \dfrac{1}{\varepsilon}\left(E_\phi(\phi) - E_\phi(v_0)\right) - O(\varepsilon) \\ \varepsilon \to 0^+, \Rightarrow \dfrac{\partial v_1}{\partial \tau} = -E_{\phi\phi}(v_0)v_1 \end{cases} \quad (27)$$

where $E_{\phi\phi}(\phi)$ denotes the Gateaux derivative (or first variation) of $E_\phi(\phi)$ [22], which is obtained based on the Mean-Value Theorem in [45]. With the initial condition $v_1(\text{x},\tau=0)=0$, we can get the unique solution of Eq. (27) as $v_1(\text{x},\tau)=0$.

Putting $v_1(\text{x},\tau)=0$ into Eq. (24) and rearranging it, we obtain

$$\frac{\partial v_2}{\partial \tau} = \frac{1}{\varepsilon}\phi_t - \frac{1}{\varepsilon^2}\frac{\partial v_0}{\partial \tau} - O(\varepsilon) \tag{28}$$

Putting Eqs. (22) and (25) into Eq. (28), we obtain

$$\frac{\partial v_2}{\partial \tau} = \Delta\phi + \frac{1}{\varepsilon^2}\left(-E_\phi(\phi) + E_\phi(v_0)\right) - O(\varepsilon) \tag{29}$$

From Eq. (23), as $\varepsilon \to 0^+$, $\phi \to v_0$, Eq. (29) can be re-written as follows

$$\frac{\partial v_2}{\partial \tau} = \Delta v_0 - E_{\phi\phi}(v_0)v_2 \tag{30}$$

with the initial condition $v_2(x,\tau = 0)=0$.

Higher order terms in Eq. (23) can be obtained by proceeding in the same way. We obtain $v_i(x, \tau)=0$, $i \geq 3$.

In summary, $\phi(x,t,\varepsilon)=v_0(x,\tau)+\varepsilon^2 v_2(x,\tau)$, where $v_0(x,\tau)$ and $v_2(x,\tau)$ satisfy the following equations, respectively

$$\begin{cases} \frac{\partial v_0}{\partial \tau} = -E_\phi(v_0) \\ v_0(x,\tau = 0) = \phi_0(x), x \in \Omega \end{cases} \text{ and } \begin{cases} \frac{\partial v_2}{\partial \tau} = \Delta v_0 - E_{\phi\phi}(v_0)v_2 \\ v_2(x,\tau = 0) = 0, x \in \Omega \end{cases}$$

**Remark A-1:** Suppose that $E(\phi)$ has $k$ local minimizers $c_1,\ldots,c_k$. Then with each $c_i$ there is an associated basin of attraction $B_i$ such that when the initial value $v_0(x,\tau=0)$ is in $B_i$, the solution $v_0(x,\tau)$ of Eq. (25) tends to $c_i$ as $\tau$ tends to infinity. Thus, $\Delta v_0$ tends to zero as $\tau$ increases, and then Eq. (30) will tend to be $\partial v_2/\partial \tau = -E_{\phi\phi}(v_0)v_2$, subject to $v_2(x,\tau = 0)=0$, whose solution is $v_2(x,\tau)=0$ as $\tau \to \infty$. We have

$$\lim_{\tau \to +\infty} \phi(x,\tau) = \lim_{\tau \to +\infty} v_0(x,\tau) + \varepsilon^2 \lim_{\tau \to +\infty} v_2(x,\tau) = c_i, \text{ if } \phi_0(x) \in B_j, j = 1,\ldots,k. \tag{31}$$

Therefore, we conclude

$$\lim_{t \to +\infty, \varepsilon \to 0^+} \phi(x,t,\varepsilon) = \sum_{i=1}^{k} c_i \chi_i(x)$$

where $\chi_i(x) \in \{0,1\}$ is the characteristic function of the set $S_i=\{x|\phi_0(x)\in B_i, i=1,\ldots,k\}$ and $B_i$ is a basin to attract $\phi(x,t,\varepsilon)$ to $c_i$. Here $c_i$ is the equilibrium solution of the LSE equation $\phi_t = -E_\phi(\phi)$ with the initialization $\phi(x,t = 0)=\phi_0(x)$. ∎

**Remark A-2:** As seen from the above proofs, we only consider the zeros of $E_\phi(\phi)$. Thus, we can readily extend the results to the case that using a function $L(\phi)$, which is not a potential, to replace $E_\phi(\phi)$, and replacing the $k$ local minima $c_1,\ldots,c_k$ of energy functional $E(\phi)$ by the zeros of $L(\phi)$. ∎